# Text to Image Synthesis using Stacked Conditional Variational Autoencoders and Conditional Generative Adversarial Networks

*Haileleol Tibebu, Aadil Malik, Varuna De Silva*
*Institute of Digital Technologies, Loughborough University, E-mail: H.Tibebu@lboro.ac.uk*

**Abstract**— Synthesizing a realistic image from textual description is a major challenge in computer vision. Current text to image synthesis approaches falls short of producing a high-resolution image that represent a text descriptor. Most existing studies rely either on Generative Adversarial Networks (GANs) or Variational Auto Encoders (VAEs). GANs has the capability to produce sharper images but lacks the diversity of outputs, whereas VAEs are good at producing a diverse range of outputs, but the images generated are often blurred. Taking into account the relative advantages of both GANs and VAEs, we proposed a new stacked Conditional VAE (CVAE) and Conditional GAN (CGAN) network architecture for synthesizing images conditioned on a text description. This study uses Conditional VAEs as an initial generator to produce a high-level sketch of the text descriptor. This high-level sketch output from first stage and a text descriptor is used as an input to the conditional GAN network. The second stage GAN produces a 256x256 high resolution image. The proposed architecture benefits from a conditioning augmentation and a residual block on the Conditional GAN network to achieve the results. Multiple experiments were conducted using CUB and Oxford-102 dataset and the result of the proposed approach is compared against state-of-the-art techniques such as StackGAN. The experiments illustrate that the proposed method generates a high-resolution image conditioned on text descriptions and yield competitive results based on Inception and Fréchet Inception Score using both datasets.

— — — — — — — — — ◆ — — — — — — — — —

## 1 INTRODUCTION

Recent advances in deep artificial neural networks (ANN)[1] and abundance of big datasets has enabled super-human performance in several practical applications such as object recognition, speech recognition and super-resolution [2]. Often referred to as deep learning or deep neural networks, utilizes multiple layers of artificial neurons to progressively extract a high level of features from a variety of data distributions [1]. The most common deep learning models used today include: Convolutional Neural Networks (CNN), Recurrent Neural Networks (RNN), Long Short-Term Memory (LSTM) and generative models such as Variational Auto-encoders (VAE) [3]–[5], Generative Adversarial Networks (GAN)[6]–[12], and Normalizing Flows (NF)[13]–[18].

Generative modelling is a branch of machine learning that tries to learn the underlying data distribution that generates data. Knowledge of underlying data distributions allows for generalization across different datasets yielding in knowledge transfer, and most importantly to deal with data-sparsity. Generative modelling has been using in multiple applications including image denoising, structured prediction, inpainting and super resolution[19]. Deep neural networks play an important role in generative modelling, by enabling learning of very high-dimensional distributions from large datasets. Prior to 2014, generative models used deep learning architectures such as Restricted Boltzmann Machines (RBM)[20], [21], and Deep Belief Networks (DBN)[22]. Though this led to positive results within the domain, a few issues arose in the training process. Training accuracy suffered from poor results due to varying data distributions and deep learning architectures that became too complex to train with[23]. As a result, Goodfellow *et al.* proposed the use of Generative Adversarial Networks (GAN) to sidestep the difficulties attained by previous generative models[24]. Through the success of this paper, research in GAN's has expanded, leading to a variety of models - including Deep Convolutional Generative Adversarial Network (DCGAN)[25], Conditional Generative Adversarial Network (CGAN)[26] and Wasserstein Generative Adversarial Network (WGAN)[27], [28]. The implementation of different variants has allowed for the development of several applications in natural language processing and computer vision, making GANs one of the most popular developments of present time. Other generative models have seen advancements including autoencoders, leading to variants such as Variational Autoencoder (VAE) and CVAE being designed with the same adversarial framework for use by deep learning practitioners.

However, despite the rapid movements in machine intelligence, the ability to capture varied high-dimensional distributions from texts, images and audio to build scalable algorithms remains a fundamental challenge in machine learning. Though the ability to mimic any distribution of data and continuous improvement of both the generator and discriminator networks presents a high potential for generative models in machine intelligence, several obstacles remain[29]. The use of GAN's is still within its infancy with variants being released for different applications with inconsistent results. This leads GANs to be highly unpredictable, with stability issues arising from training and difficulty to train. Concerns arise with its effectiveness being varied from application to application. On the other hand, the use of VAEs suffer from getting the model to learn meaningful features and for the latent space to present a generalized stable representation. As promising as the new generation of generative models are, training and producing plausible results are sparse, with more research needed to discover training tricks or hints for the model to perform successfully[30].

Generating and image from a text description, referred to as Text to image synthesis, is regarded as one of the most elusive goals for current computer vision research. This is because, if an algorithm can generate realistic images from text descriptors, then we can have high confidence that the algorithm understands the features in the images. Despite the challenges of using GANs and autoencoders, success has been found in performing text to image processing. RNN was initially used for text conversion, which uses natural language for describing the space in the image's visual vicinity[31]. Using this network, the distribution of images was labelled within texts as multimodal models. This presented a natural use of generative models, with GANs ability to predict real images from fake images in its adversarial play. This allows the images generated to be more accurate. Reed *et al.* was the first to investigate the possibility of training GAN's for image synthesis from a list of labels[32]. He found success in this endeavour and since then, a few papers have attempted to better the results using the same fundamentals of GANs. StackGAN forms 2 stacks of GAN networks to synthesize images from sentences. Through this method, the images were sharper and of higher resolution, accurately reflecting the sentences. However, this model is difficult to replicate as it can collapse, leading to training becoming predictable[33]. To address some of the issues of this initial paper, StackGAN++ [34] was used with a different methodology to produce greater results. Although there is proven success with GANs, research into other generative models have shown similar success as well. Bian *et al.* used a combination of variational autoencoders and generative adversarial models for his generator to produce synthesized images[35]. Nguyen *et al.*'s research looked at using iterative activation function within his generative model to generate images[36]. Recent studies also show significant success in face manipulation tasks using the advanced GANs and VAEs[37][38]–[41]

GANs have a relative advantage of generating data which is similar to the real data because they don't assume any specific probability density estimation. The advantage of VAE is that it follows a specific probability distribution (Gaussian distribution) which allows a model to learn smooth latent state representation of input data.

To address the issues arising from the relative disadvantages of GANs and VAEs, this paper proposes the use of conditional versions of GANs and VAEs as a stacked architecture. Our hypothesis is that ability of GANs to produce sharp images can be coupled with the VAE's ability to learn more generalizable distributions will yield sharper images across a broad range of textual definitions. Furthermore, such a stacked architecture allows for modular training of these networks allowing for more stability during the training process.

The rest of the paper is organized as follows. A review of recent significant progress achieved towards text to image synthesis is presented in the section 2. Section 3, defines and discuss the necessary background knowledge. We then present the proposed method architecture followed by evaluation of method performance in section 4. The discussion of results produced by the proposed method is presented in section 5, followed on by the conclusion.

## 2 RELATED WORK

Text to image synthesis is an attractive research subject, which has received much attention as a holy grail venture that attempts to combine representation learning in natural language processing (NLP) and computer vision.

The first implementation of using a GAN model was by Reed et al [32], used a conditional-like GAN based on convolutional networks for both the generator and the discriminator. Similar to a conditional GAN, a conditioned vector of the embedded text descriptors is concatenated with the GAN as opposed to traditional methods of labels/attributes. The convolutional networks are based on the same architecture as a DCGAN [25] with the results showing a plausible generation of 64*64 images of birds and flowers. Though successful, the results lacked details of lurid features such as the wings of a bird or the shape of a petal for example. In addition, the generated images are unable to synthesize to higher resolutions without additional objects. As a result, the generator failed to detangle different regions of the image, such as the background, contributing to a lack of clarity in the generated images[42].

Building upon Reed S. et al's [32] work, StackGAN [33] proposes using two GAN generators for synthesising images. The first set of images generated from the first generator are of low-resolution quality (64*64), that contain a rough alignment of the shapes and colours of the objects. These images are then fed into a second generator which produces results of a higher resolution and sharper details at 256*256. Both generators use conditional GANs with a text embedding. Reed et al. [32] encountered a problem where the text embedding is limited to only descriptors of the images and not sentences. To mitigate the problem, Zhang et al. [33] use a conditioning augmentation which allows the latent variables to randomly sample from an independent gaussian distribution. To ensure the smoothness of the conditioned embedding, a Kullback-Leibler divergence regularization term is added to the generator as it is trained. The performance of the generator was considerably improved from his initial paper, with increased quality and sharpness, which is reflected through higher scores that are used to measure GAN performance. However, there were varying degrees of clarity between images and a lack of focus in the background images.

To further improve on StackGAN, StackGAN++ [34] uses an additional pair of generators and discriminators to improve the sharpness of the image. To ensure consistency, a regularization term calculated as the difference between the mean-square loss and the variance between the generator and discriminator is used. Though this addressed the varying degree of clarity in the previous iteration, the lack of sharpness in the background of the generated images has still not been accurately addressed[43].

Aside from Reed et al's. [32] approach to image generation, another iterative approach developed called "Plug and Play Generative Networks" (PPGN) proposed using activation maximisation to generate images[36]. PPGN's are composed of an image generator that is a combination of denoising autoencoder, a GAN and a pre-trained image captioning model, which introduces an additional prior on the latent code. The results generated higher resolution images with higher quality than Reed et al. [32] and Zhang et al. [33] models. Though the iterative sampling method model takes a considerable amount of time to generate, its performance is amongst the best for text to image synthesis. However, the model does suffer from mode collapse by the generator, where the images generated are of limited varieties at random points during training. Though steps were taken to mitigate the risks, such as batch normalization which allows a variety of data to be trained, the problem remains when potentially training the network and no permanent solutions have been found [44]. Another methodology looks at combining the use of two generative models with VAE and GAN. Bian et al. [35] combine the use of an encoder network, generative network, discriminative network and a classification network to generate images. These images are generated without text descriptions using asymmetric loss functions which models an image as a composition of label and latent attributes in a probabilistic model. This model achieved a varying degree of results. [45] Also use VAE for context aware image generation mainly to separate the foreground and background of an image then uses GANs to refine the results from VAE.

Present text-to image models all suffer from several limitations. Currently, the models perform well on single object images, such as birds in CUB, flowers in Oxford-102 and individual objects in ImageNet. However, the models severely underperform where multiple objects are involved, such as the MS-COCO dataset. This is likely because the models are unable to learn the features of each object but only knows the rough shapes and images between them [46]. In addition, training text to image models takes a considerable amount of time due to the large training sets and a novel step is needed to teach the concept of objects in everyday datasets. Currently, the commonly used datasets are a specific, well-defined set used for concept and theory testing [43].

## 3 PRELIMINARIES

The purpose of generative modelling is to learn an underlying probability distribution $P(X)$ by leveraging the observed data $X$. Often this process involves, the selection of a suitable probability density function parameterized by certain parameters ($\theta$), and using a method such as Maximum Likelihood or Maximum A Posterior estimation to estimate the parameters. However, as the data becomes high dimensional, such as images, or text, selecting a distribution, and finding the parameters becomes a non-trivial task. Latent variable models are a method to learn the underlying distribution $P(X)$ as a mixture of distributions over hidden variables ($z$). Variational inference is a popular method to approximate the complex distribution over z, with a known family of distributions.

With the emergence of Big data, and the neural networks as complex function representations, the methods which leverage data to learn the parameters of distribution has become popular.

Variational Autoencoder (VAE) is a neural network based generative model that aims to capture data distributions such as images, audio, video through variational inference [47]. A VAE bears close structural resemblance to an autoencoder. It is made up of an encoder known as the inference model and a decoder which acts as the generative model. The VAE constructs its input data while learning through a continuous latent vector. Due to the focus of VAE being on variational inference, it is an efficient model for Bayesian inference with latent variables [48]. The encoder $Q_\phi(z|X)$ encodes the data instance $x$ into a latent representation space $z$, for learning the distribution of hidden variable with a Kullback-Leibler (KL) divergence penalty. The decoder $P_\theta(x|z)$ decodes $z$ back into the original data space by minimizing the reconstruction error. The encoder and decoder is implemented as a neural network of parameters $\phi$ and $\theta$ respectively.

Therefore, the objective of the VAE can be demarcated in Eq.1 as:

$$\log P(X) - D_{KL}[Q(z|X)||P(z|X)] = E[\log P(X|z)] - D_{KL}[Q(z|X)||P(z) \qquad (1)$$

where $D_{KL}$ denotes the KL divergence, and the log-likelihood $\log P(X|z)$ is converted into its lower bound by variational inference for an efficient solution.

In VAE the encoder models the latent variable $z$ based on $X$ without giving much care about the different type of $X$. For instance, it does not much attention for the label of $X$. This is also similar in the decoder part. To improve this weakness of VAE, we apply conditioning technique the encoder and decoder as stated in [33].

A conditional VAE is an extension of VAEs, where the algorithm models the latent variables and data, conditioned on certain random variables. The encoder and the decoder both has the conditioning variable as an input [49]. The objective of a conditional VAE is given as,

$$\log P(X|c) - D_{KL}[Q(z|X,c) \| P(z|X,c)] = E[\log P(X|z,c)] - D_{KL}[Q(z|X,c) \| P(z|c)] \qquad (2)$$

The difference between Eq 1 and Eq 2 can be expressed as, all the distributions in Eq. 2 are conditioned with a variable c.

A generative adversarial network (GAN) is a deep neural network architecture introduced by Ian

Goodfellow *et al.* [24] as a step-forward on existing generative models for both unsupervised and semi-supervised learning. Made up of a generator and discriminator, both neural networks in a GAN are trained through adversarial play, where both networks compete against each other. Through multiple iterations of generation and discrimination, both networks attempt to outwit one another, whilst training each other. The purpose of the generator $G$ is to generative new data $G(z)$ from existing data through a randomly generated latent space $z$. The discriminator $D$ attempts to differentiate between the real data and the fake data generated from the generator[50].

The objective for the GAN is to learn a distribution where $P(x)$ matches $Pz(z)$ over data x via a min-max game. This state is known as the Nash equilibrium in game theory. The objective function is shown in Eq. 3.

$$\min_G \max_D V(D, G) = E_{x \sim pdata(x)}[\log D(x)] + E_{z \sim p_z(z)}[\log (1 - D(G(z)))] \qquad (3)$$

Where the D(x) is the discriminator model, G(z) is the generator model, $P_x$ is the real data distribution, $P_z$ is the data distribution generated by the generator and E is the expected output.

CGAN is a deep learning methodology that places a conditional setting on the network, whereby the generator and discriminator are conditioned on auxiliary information such as class labels or data from other modalities [51]. A conditional input $y$ is concatenated with random noise z so that the image that is generated is defined by $G(y, z)$. As a result, the objective function of the min-max game would be updated as shown in Eq. 4.

$$\min_G \max_D V(D, G) = E_{x \sim pdata(x)}[\log D(x|y)] + E_{z \sim p_z(z)}[\log (1 - D(G(z|y)))] \qquad (4)$$

Where the *D(x)* is the discriminator model, *G(z)* is the generator model, $P_x$ is the real data distribution, $P_z$ is the data distribution generated by the generator and *E* is the expected output.

## 4 THE PROPOSED MODEL ARCHITECTURE

The aim of proposed method is to design a new generative neural network architecture that can generate images conditioned on a text caption. In this section we present the methodology based on conditional generative models discussed in section 3, and propose a stacked generative neural network architecture based on conditional GANs and conditional VAEs to accomplish this objective.

### 4.1 Conditioning Augmentation

When limited number of training text-image pairs data feed to the generative models, the network experiences sparsity in the text conditioning manifold which makes it challenging to train GAN. To overcome this, Zang et al. [33] proposed a conditioning augmentation approach which promote smoothness in the latent conditioning manifold. The method allows few random perturbations in the conditioning manifold. It also increases the diversity of synthesized images.

#### 4.1.1 The stacked conditional generative neural network architecture

The proposed model will be split into two stages. In the first stage, a Conditional VAE is used to generate 64x64 low-resolution images based on the dataset and the given text descriptions. These low-resolution images will be fed through an encoder and a decoder with the output containing the basic shapes of the objects, distorted colours and rough sketching to match the text descriptions as the neural network learns how to accurately synthesise images. In the second stage a conditional GAN is trained to generate 256x256 higher resolution images, taking the input from the decoded low-resolution images from stage 1. In addition, text embeddings will be used to allow the model to condition on the given input. The output will be sharper, brighter images that will accurately reflect the text descriptions with photo-realistic results. Through continuous adversarial play, the images will become sharper until they look photo-realistic and closely match the text descriptions. The proposed architecture is illustrated in Figure 1.

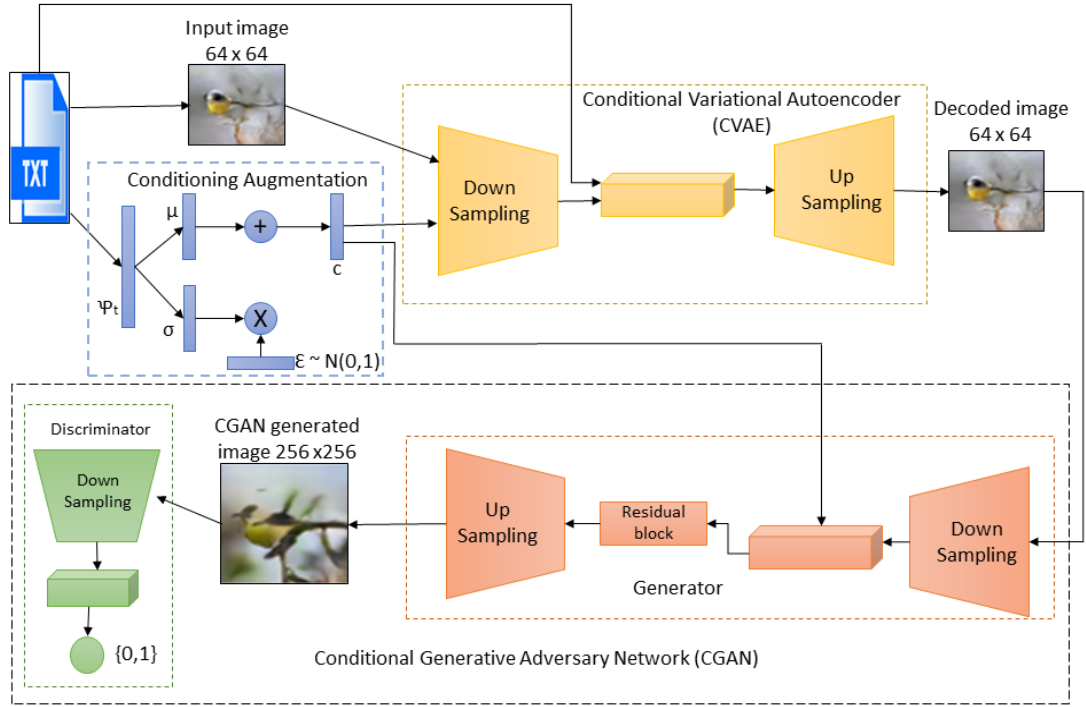

*Figure 1: Proposed CVAE-CGAN Stacked Model*

### 4.2 Stage I: Conditional VAE (C-VAE)

Objective of the conditional VAE is to extract key background and object details to synthesise high-resolution images. The CVAE model's utilization from prior research has been shown to be excellent at detecting the backgrounds of image shapes and colours.

The encoder will take an image and its associated text embeddings as the input. To allow for the images and the text embeddings to be read by the encoder together, they need the same vector shape. Therefore, the text embedding is flattened using a dense layer and reshaped with another layer. Once these inputs have the same base vector size, their strings will be concatenated together and merged into one input entity. Additionally, the text embeddings are merged into a conditioning augmentation. This method is utilized to allow the model to learn representations from the images for the latent vector. The encoder consists of a 5x5 convolutional stride 2 architecture, to extract the key features of the images and texts until there is an output dimension of 2048. This output is formatted as the lambda variational log weight of the images which will be fed into the decoder. The encoder uses batch normalization and ReLU activation. This allows for smoother backpropagation and smoother computation of the model, as discussed in the literature review. An overview of the encoder is shown in
Figure 2.

When decoding the image, the decoder takes the encoded lambda variational log weight of the image and concatenates this with the text embedding, which is placed in a dense layer for it to be reshaped with the same baseline vector. To decode, the decoder simply does the opposite for the encoder with 5×5 fractional upscaling of the images via transposing the convolutional direction. This is also done with stride 2 until it generates a 64×64 image. Each of these convolutional layers is placed with a layer of batch normalization and ReLU activation functions until the final layer. The final layer contains a Tanh activation function without batch normalization. Tanh is used for the classification as it allows for binary values between 1 and -1. An illustration of the decoder is shown in Figure 2(B). The loss function of conditional VAE is defined in Eq. (5).

$$-E_{z \sim q(Z|X)}[\log p(X|Z,C)] + KL(q_{\emptyset}(Z|X,C)||p(Z,C)) \qquad (5)$$

The first term of the loss in Eq. 6 is the reconstruction error. In other words, it is expected negative log likelihood of the given data point. Large error means the decoder is not constructing the data efficiently. The second term in Eq (5). Is Killback-Leibler divergence between $q_{\emptyset}(z|x,c)$, encoding distribution, and p(z,c). ∅ is the variational parameter.

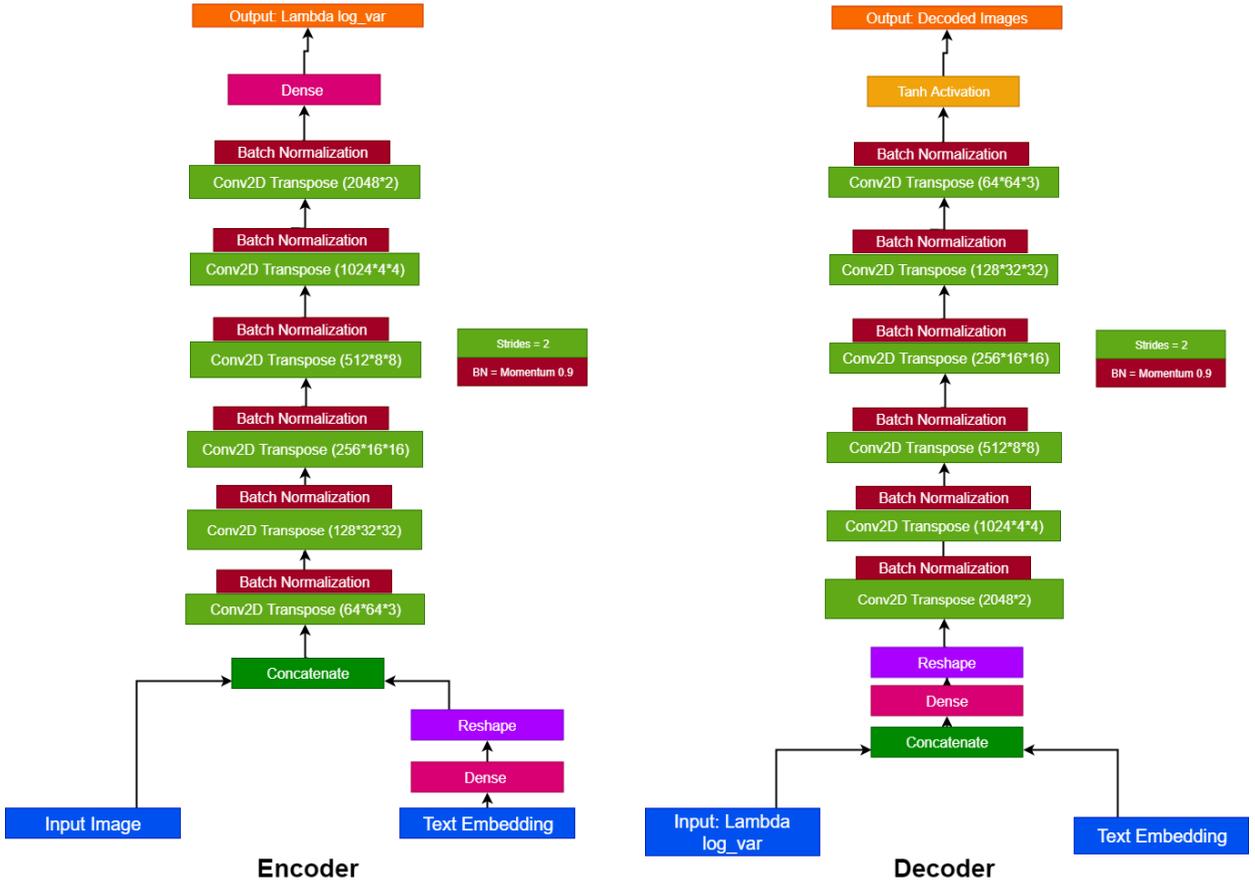

*Figure 2 (A) Proposed Encoder Network Architecture for C-VAE, (B) Proposed Decoder Network Architecture for C-VAE*

### 4.3 Stage II: Conditional Generative Adversarial Network (C-GAN)

The decoded image from the C-VAE module is used as an input into the generator of the C-GAN. The low-resolution images currently lack sharpness and clarity. The C-GAN will use multiple layers of up-sampling and down-sampling convolutional layers to allow for key extractions from the image and text, to improve the resolution and the details of the low-resolution output of the C-VAE stage.

The generator takes the low-resolution images and places it through three layers of down-sampling convolutions stride 2. This is then coupled with the lambda value of the text embedding to allow for further learning. Following this, the generator uses 5×5 stride convolutions within the architecture that upsizes and downsizes the image to extract key qualities from both the text embedding and the images being fed to the model. Since GANs suffer from mode collapse and vanishing gradients, this research uses batch normalization and ReLU activation to stabilizes the model and ensure the discriminator receives meaningful images. This model again uses tanh activation function as a final layer and generated images are produced of 256×256 photo-realistic quality. An illustration of the generator is shown in Figure 3 (A).

The image generated from the generator is then fed directly into the discriminator network. Using 5x5 convolutions with stride 2, the image is downscaled until it reaches 2048 dimensions. This is then downscaled to 512 dimensions until it is put through a flatten and dense layer for the discriminator to predict if it is a real or a fake image. In the discriminator, the images are placed through more convolutions to allow the network to break down the representations of the image so that it can adequately learn whether this image is real or synthetic. This will allow the adversarial play to lengthen itself as the discriminator will take in more representations than the generator produced the image with.

This, in turn, allows the discriminator to learn at a gradual rate, which allows the images generated to be of higher quality. To allow the discriminator to stabilize, this research uses batch normalization and leaky ReLU. Leaky ReLU is used as opposed to ReLU. This is because, as the images have now gone through three different networks, some of the gradients may be fragile. These may not activate given the number of hidden layers that the inputs have gone through, including the encoder, decoder and the generator. This could mean that important gradients that are small in numeric value may be taken by the network as 0 from the initial layers. To combat this, Leaky ReLU allows a small gradient to be placed, with a small value. For this research, the alpha value is equal to 0.2, to ensure the model utilizes the gradients from the earlier layers. Figure 3 B) shows the breakdown of the discriminator. The CGAN used in this paper train the generator $G$ and the discriminator $D$ by maximising $\mathcal{L}_D$ in Eq. (6) and minimising the $\mathcal{L}_G$ in Eq. (7).

$$\mathcal{L}_D = E_{(I,t) \sim pdata}[\log D(I, \phi t)] + E_{s0 \sim pG0, t \sim pdata}[\log(1 - D(G(s0, \hat{c}), \phi t))], \quad (6)$$

$$\mathcal{L}_G = E_{s0 \sim pG0, t \sim pdata}[\log(1 - D(G(s0, \hat{c}), \phi t))] + \lambda D_{KL}(N(\mu(\phi t), \Sigma(\phi t)) \| N(0, I)) \quad (7)$$

Where the real image $I$ and description of the txt $t$ are form the data distribution $pdata$. $z$ is a vector of noise which randomly sampled from a $p_z$ gaussian distribution. The regularization parameter $\lambda$ balance the two terms in Eq. (7). The architecture of the generator and the discriminator is presented in fig 3.

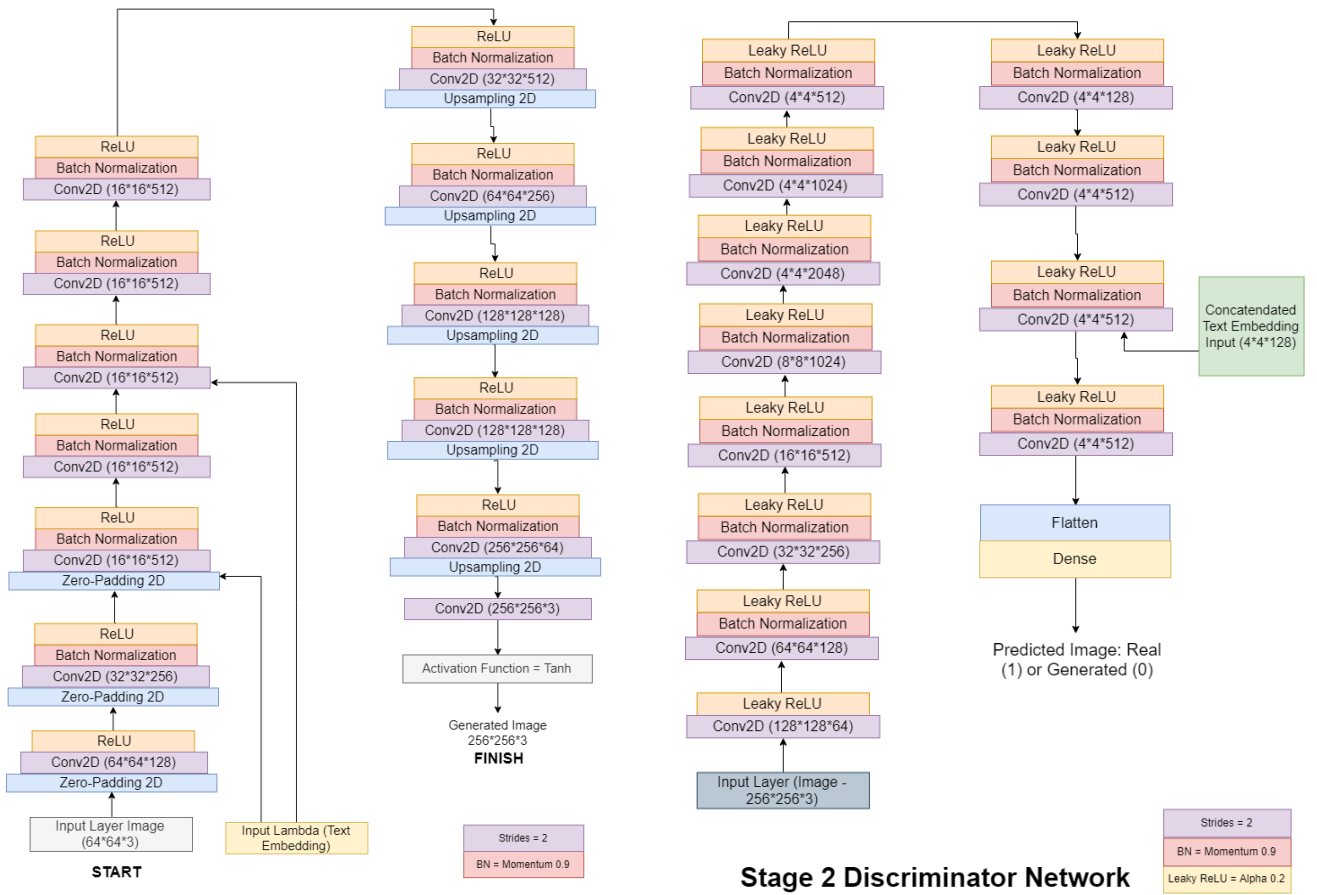

*Figure 3: (A) Proposed Generator Network Architecture, (B) Proposed Discriminator Network Architecture*

Through adversarial play, the quality of images will improve with each epoch, until the discriminator and the generator converge. When convergence occurs, this indicates that the model has reached its optimum point.

### 4.3 Hyper-parameter optimization

This study takes a sample noise from a Gaussian distribution where the noise vector dimension is 100. This is then concatenated with the text embedding. The C-VAE model is trained for 150 epochs for the first stage of the model. Afterwards, the model is trained further for 150 epochs in the second stage. The training process is powered by Adam optimizer with an initial learning rate of 0.0002 for the CVAE which is decayed by a factor of 0.2 every 25 epochs. For the CGAN, a learning rate for both the generator and discriminator is set as 0.002 with the same decaying rate of 0.2 every 25 epochs. The defining variables and hyperparameters are presented in Table 1.

Table 1: Hyper-parameter values used in the NNs

| Hyperparameter | Value |
| --- | --- |
| **Noise vector** | 100 |
| **Conditional vector** | 128 |
| **Batch size** | 64 |
| **Learning rate (Stage 1)** | 0.0002 |
| **Learning rate (Stage 2)** | 0.002 |
| **Epochs (Stage 1)** | 150 |
| **Epochs (Stage 2)** | 150 |
| **Decay Rate** | 0.2 every 25 epochs |

### 4.4 Datasets and Pre-processing

For the model is trained and evaluated on two publicly available datasets. These two datasets are widely used by the community and have been the data input for previous text to image synthesis research [31][32][52][53]. This allows for comparative analysis of results to evaluate the effectiveness of the proposed model. The CUB dataset [54] contains 200 bird species with 11,788 images. Over 75% of the images in this dataset have object-to-image size ratios of less than 0.5. As a pre-processing step, the model crops all images to ensure that bounding boxes of birds have greater than 0.75 object-image size ratios. 20 captions are used for the images in the dataset provided by Reed *et al*. [9] are utilized in the paper. The CUB dataset split into 150 training categories and 50 testing categories. These were split as class-disjoint training and test sets. The second dataset utilized is the Oxford-102 flower set [55] which contains 8189 flower images of 102 categories. This is split into 82 training categories and 20 testing categories.

### 5 EVALUATION OF MODEL PERFORMANCE

Evaluating the performance of generative models (GANs and VAEs) is a difficult process, because there is no single metric that can judge the quality between synthetic and real images accurately enough. Consistent with related literature, this paper will use the following methods to evaluate how well the model produces images.

**Inception score:** this is a common method used by similar papers [32]–[34], which is used to quantitatively summarize the quality of the generated images. The inception score is calculated by using many generated images, which are classified using the model. Specifically, the probability of the image belonging to each class is predicted. These probabilities are summarized to score how the image belongs to each class and t he diversity of the images. The higher inception score indicates the better – quality generated images.

**Fréchet Inception Distance (FID)**: this method is a recently proposed metric that uses an inception check to extract features from an intermediate layer of the images. This is then modelled as a data distribution, with the extraction of features using a multi-variate Gaussian distribution with the mean and covariance. This is calculated for both the synthetic and real images, with the distance between them as the score. The lower the score of the FID, the closer distance between the synthetic and real data distributions. Therefore, lower FID values mean better image quality and diversity.

**Generator and Discriminator losses**: This evaluation method will look at the discriminator and generator losses that come from the final C-GAN stage of the model. The trend line for each network will allow to observe whether the model would continue to learn, subject to more training. If the discriminator values remain slowly falling and not towards 0, the discriminator is continuing to learn. In addition, the convergence is a sign of the model continuing to learn and the potential of this methodology given its promise.

Subjective Visual Comparisons – This method will involve the practitioner looking at a sample of the generated images and how well they are computed and understood by the network compared to other approaches [32]–[34]. This will be assessed by this researcher of this project.

| Dataset | Method | Inception Score | Fréchet Inception Score |
|---|---|---|---|
| CUB | GAN-INT-CLS | 2.88 ± 0.04 | 68.79 |
|  | StackGAN | 3.70 ± 0.04 | 51.89 |
|  | StackGAN++ | 4.04 ± 0.05 | 15.30 |
|  | **OURS** | **1.94 ± 0.02** | **74.12** |
| Oxford-102 | GAN-INT-CLS | 2.26 ± 0.03 | 79.55 |
|  | StackGAN | 3.20 ± 0.01 | 55.28 |
|  | StackGAN++ | 3.26 ± 0.01 | 48.68 |
|  | **OURS** | **2.61 ± 0.06** | **65.13** |

Table 2: IS and FID Scores for the images generated

### 5.1 Experimental Results

Firstly, the proposed approach will be compared to three text to image generation methods that have made excellent strides within the domain on the CUB and Oxford 102 datasets. These are namely the StackGAN [33], StackGAN++ [34] and the GAN-INT-CLS [32] approaches. They will be evaluated using four metrics mentioned in the previous section

#### 5.1.1 Inception Score and FID Results

The results shown in Table 2 show that this model achieved the lowest inception score and fréchet inception score on the CUB dataset. It had a 50% lower value on the inception score (1.94 to 2.88) and a 7% higher distance on the FID score (74.12 to 68.79). However, on the Oxford-102 dataset, the CVAE-CGAN model ranks higher than GAN-INT-CLS, though both failed in attaining a higher score than the StackGAN and StackGAN++ approaches. Since FID is more robust to potential noise, it is better for image diversity and this is shown in examples generated by the model. On the Oxford-102 dataset, this approach showed a 16% improvement on the inception score and an 18% improvement on the Fréchet inception distance compare to the GAN-INT-CLS approach.

#### 5.1.2 GAN Performance

The results for generator and discriminator performance for CUB dataset and the Oxford-102 dataset is shown in **Error! Reference source not found.** (A) and **Error! Reference source not found.** (B) respectively. For the CUB dataset for which results are illustrated in figure 4 (A), the networks still have the capability to learn further. However, the fact that the discriminator holds relatively steady, shows that it can't distinguish between real images and fake images generated from the generator. In addition, the generator is not able to create a synthetic image perfectly that is able to fool the discriminator. While the convergence has not been achieved and the Nash equilibrium is still not accomplished, it is closer towards achieving this. For the Oxford-102 dataset, the model hasn't achieved convergence but is closer towards it with the discriminator slowly falling towards 0. This is reflected in the images generated, as the images were comparable to other models. The potential of the training further the model will be discussed in the next section.

### 5.1.3 Visual Comparison Results

Subjective visual comparisons are illustrated in Figure 4. In line with the IS and FID results, the images generated from the Oxford -102 dataset is more accurately reflected than with the CUB dataset. In particular, the 3rd and 4th columns produced excellent results where sharp colours and photorealism that surpasses the StackGAN and StackGAN++ models. For the 2nd column, the colour is inconsistent with the text description with a different shade of the colour compared to other models. In short, the results for the Oxford – 102 as a visual comparison was positive with generating text-captioned images.

In respect to the CUB dataset, the CVAE-CGAN model performed poorly in quantitative results. Subject to visual comparisons, the user can see that the results generated for the first three columns depict the right photo, but without the clarity and sharpness that was seen in the Oxford-102 dataset. The 1st column, though following the text description, lacks the fine-grained quality of the StackGAN and StackGAN++ model. In the 2nd column, the results look more accurate than any other model without the sharpness and clarity. All the text descriptions are fulfilled without it. Fig 4 presents the visual results in comparison with similar algorithms. For the last two columns, it seems as though the training of the model had trouble with the fine differentiation of darker colours. This has led to colour distortions in the images as shown. The potential reasons why the results were mixed will be discussed in the next section.

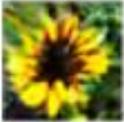

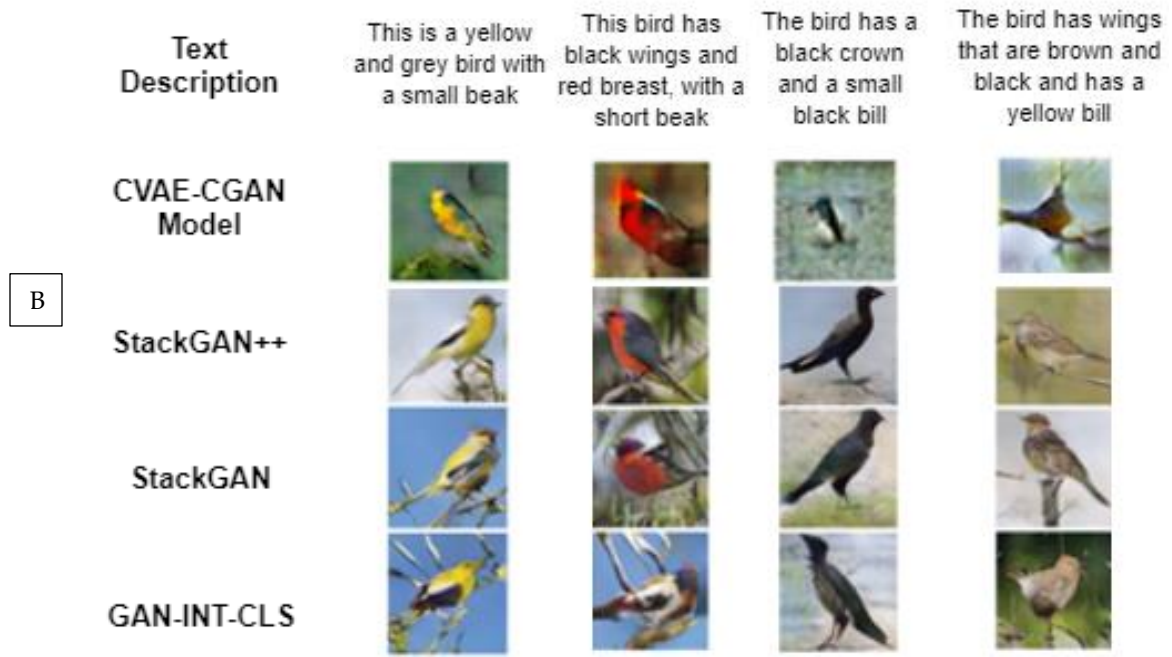

Figure 4: Visual Results of all the models compared in the study, (A) Samples from the Oxford-102 data set (B) Samples from the CUB data set

## 5 DISCUSSION OF RESULTS

The proposed CVAE-GAN approach performed better on the Oxford-102 dataset compared to the CUB dataset according to the Fréchet inception distance and inception score evaluation metrics. The subjective visual comparisons show a similar trend to the quantitative results. Considering the Oxford-102 dataset examples, the proposed method illustrates excellent results with better clarity than all other approaches. In comparison the performance on CUB dataset shows mixed results, where the StackGAN method performs better. In this section we shall discuss these results and their implications in detail.

The model was trained for 300 epochs. Overfitting was not a problem given that the mode did not collapse as per the image quality and that the bias/variance trade-off was mitigated using the methods mentioned by Salimans T.[30], such as batch normalization and historical averaging. It is entirely possible that if the model was run further than 300 epochs, that it would collapse and become unstable to generate images. Therefore, further training may be required to optimize the hyperparameters for both stages of the framework.

The present evaluation metrics used for measuring GAN performance in computer vision is severely limited. The use of FID and IS look at the quality of the images produced, and the diversity of the images generated. Since both measures are based on feature extraction, being the presence or absence of features, the measures suffer heavily from potential mode collapse. Given that the images generated are only from 8 text descriptors between both datasets, this can lead to some high bias and low variance. Though this was mitigated using conditional GAN which mixes classes data over one-class data, we found that despite the good results for the Oxford-102 dataset, visually poor images in their evaluation did well with the FID results.

Currently, the present text to image approaches works well on datasets which contain single objects per image, such as birds for CUB [27], flowers for Oxford-102 [28]. However, when one uses image sets such as MS-COCO [29], the text to image domain suffers from poor results. Multiple objects within the foreground or background of the images lead to suffering in the images generated. This is perhaps the reason why this approach achieved significantly better results on the flowers dataset than the bird's dataset. The bird's dataset had different angles of the birds with a variety of different backgrounds and other objects in the dataset such as trees, skies, branches, clouds and other small objects which may have been inaccurately picked up from the model as a significant feature. In contrast, the flowers dataset had images of the flowers that focused on the flower itself. The features that needed to be extracted for birds is significantly more precise than for flowers and may lend reason as to why this research's approach performed well.

The breaking down of the image generation into two stages allows each network to focus on the finer detail of creating images. This can be expanded further with three or more stacks of generative models to allow for the networks of generating high-resolution images. This can be achieved by potentially having one stage focus on the background of the images, one on the foreground and the other model focusing on the upscaling of the image quality. This can be done through more stable generative model variants such as DCGAN [32], which excel at image synthesis. Though stabilizing the model will be a challenge, there is potential for producing greater results and merits further research.

Although significant progress has been achieved for image generation with GANs, the challenge remains of accomplishing greater efficiency in aligning input text with a generated image. One of the problems that practitioners suffer from include, GANs not being a suitable model to work with for natural language processing compared to Recurrent Neural Networks (RNN) & Long Short-Term Memory (LSTM). GANs excel more towards computer vision. RNNs can be used to model a sequence of the text input, with each layer dependent on one another. This allows it to be used with convolutional layers. LSTMs can be used in this instance as well, for learning longer-term text descriptions.

One area of development to consider is using RNN and LSTMs to break down the input text for the generative model. For instance, the use of RNN and LSTMs would allow for more descriptors in the text to be used. These descriptors will serve as features, which can be embedded within the text. This text would then be shared as a layer between images using conditioning augmentation. This is potentially one route that can be taken with utilizing other areas of artificial neural networks, to allow for the algorithm to have a full understanding of its task [51].

As mentioned in this research's results, text to image synthesis in its current state can accurately and capably produce images on datasets with single object per image. When there are more objects in an image, the results begin to vary within the domain. An area to tackle through research as previously discussed, is finding a novel way for the algorithm to learn the features of each object rather than learning the concept. This can be potentially achieved by training separate models that can learn and create individual objects. Another model would then be created, with the task of learning how to combine these objects into an image based on text descriptors. This would allow for the image to establish a relationship between the objects and learn the features in an efficient break down. This is opposed to learning the concept of the design through complicated images which contain numerous objects [16]. However, this would require substantial training datasets for different objects with numerous iterations of pre-processing for model utilization. The same would apply for a larger dataset of images containing the objects together. As of writing this paper, it is extremely resource consuming to create or acquire such a data set.

## 6 CONCLUSIONS

This paper studied the use of a stacked generative model of Conditional Variational Autoencoders (CVAEs) and Conditional Generative Adversarial Networks (CGANs) for producing an image conditioned on a text descriptor. The proposed methodology is formulated as two stage process that makes the training of the neural network architecture more stable. The first stage contains the CVAE, which focuses on extracting and producing the shapes of the objects and their colour. The aim of the CVAE is to output low-resolution images (64x64), which match text descriptions to generated images. The second stage contains the CGAN, which takes the images generated from the first stage and refines those images, improving the quality of those images. This is done by down-sampling and up-sampling the images through convolutional layers. The second stage generates sharper and brighter images of resolution 256×256. The proposed approach was tested on two publicly available datasets. The proposed architecture illustrated competitive performance on the Oxford-102 and CUB dataset when compared with competing architectures such as StackGAN and StackGAN++. The future work will involve the study of advanced VAE and GAN architectures, in combination with symbolic representation mechanisms to develop explainable, and transferable generative models.

**Funding:** This research was funded by EPSRC, grant number 2126550 510

**Acknowledgements:** This work was supported by EP/T000783/1: MIMIc: Multimodal ImitationLearning in MultI-Agent En-511 vironments and EP/S515140/1: Intuitive Learning: new paradigm in AI for decision making in intelligent mobility.